\documentclass[letterpaper, 10 pt, conference]{ieeeconf}  %

\IEEEoverridecommandlockouts                              %

\author{Erencem Ozbey$^{1,*}$, Fethiye Irmak Do\u{g}an$^{2,*}$, Jin Huang$^{2,*,\dagger}$, and Hatice Gunes$^{2}$%
\thanks{$^{1}$E. Ozbey is with Bogazici University, Istanbul, Turkiye
        {\tt\small erencem.ozbey@std.bogazici.edu.tr}}%
\thanks{$^{2}$F. I. Do\u{g}an, J. Huang, and H. Gunes are with the University of Cambridge, Cambridge, United Kingdom
        {\tt\small \{fid21, jh2642, hg410\}@cam.ac.uk}}%
\thanks{$^{*}$These authors contributed equally and are co-first authors.}%
\thanks{$^{\dagger}$Corresponding author.}%
\thanks{This work was mainly carried out while E. Ozbey was a visiting undergraduate student at the AFAR Lab, Dept. of CST, University of Cambridge. The works of F. I. Do\u{g}an \& H. Gunes were supported in part by Google via a Google Initiated Grant (GIG) and by CHANSE \& NORFACE through the MICRO project, funded by ESRC/UKRI grant ref.~UKRI572. J. Huang's work was supported by the EU's Horizon Europe research and innovation programme under the Marie Sk\l{}odowska-Curie Actions Postdoctoral Fellowships (European Fellowship) 2024, grant agreement no.~101203728 --- SOCIALADAPT --- HORIZON-MSCA-2024-PF-01. Views and opinions expressed are those of the author(s) only and do not necessarily reflect those of the funding bodies. Neither the EU nor the granting authorities can be held responsible. %
\textbf{Open access:} For the purpose of open access, the authors have applied a Creative Commons Attribution (CC BY) license to any Accepted Manuscript version arising. 
\textbf{Contributions:} Conceptualization: JH, FID, HG. Funding acquisition: JH, HG. Methodology: JH, FID, EO. Investigation: EO. Writing: FID, JH, EO. Supervision \& project administration: FID, JH, HG.}%
}

\title{\LARGE \bf
StARS: Socially Appropriate Robot Actions via a Recommender System-Driven Approach %
}

\usepackage{xcolor}
\usepackage{graphicx}
\usepackage{subcaption}
\usepackage{bm}
\usepackage{bbm}
\usepackage{dsfont}
\usepackage{xspace}
\usepackage{url}
\usepackage{multirow}
\usepackage{booktabs}
\usepackage{enumerate}
\usepackage{acronym}
\usepackage{balance}
\usepackage{mleftright}
\usepackage{float}
\usepackage{tikz}
\usepackage{cite}
\usepackage[hidelinks]{hyperref}
\usepackage{algorithm}
\usepackage{algpseudocode}
\algnewcommand{\LineComment}[1]{\State \(\triangleright\) #1}

\usepackage{amssymb}
\usepackage{amsmath}
\usepackage{comment}
\usepackage{adjustbox}
\usepackage{graphicx}
\usepackage{subcaption}

\newcommand{\eg}{\emph{e.g.,}\xspace}

\newcommand{\ignore}[1]{}

\definecolor{PreferredGreen}{RGB}{34,139,34}

\acrodef{RS}{recommender system}
\acrodef{NLP}{natural language processing}
\acrodef{ML}{machine learning}
\acrodef{CV}{computer vision}
\acrodef{LLM}{large language model}
\acrodef{VLM}{vision language model}
\acrodef{VLAM}{vision-language-action model}
\acrodef{CF}{collaborative filtering}
\acrodef{CB}{content-based}

\acrodef{HRI}{human robot interaction}
\acrodef{HCI}{human computer interaction}
\acrodef{MF}{matrix factorization}

\begin{document}

\maketitle
\thispagestyle{empty}
\pagestyle{empty}

\bstctlcite{IEEEexample:BSTcontrol}

\begin{abstract}

Social appropriateness in human-robot interaction (HRI) is not universal: different people can judge the same robot action differently in the same situation. To capture this inter-subject variability, we reformulate socially appropriate action generation as a preference modelling problem inspired by recommender systems, treating annotators as \textit{users}, contexts/scenes as \textit{items}, and appropriateness scores over a set of candidate robot actions as \textit{targets}. We propose StARS, a novel model-agnostic framework that integrates collaborative filtering with learnable scene representations to generate user-specific appropriateness scores over candidate robot actions.  StARS is \emph{model-agnostic}: it can be integrated with various scene encoders and backbones, enabling personalisation without redesigning the underlying model. We evaluate StARS on two socially aware robotics datasets, MannersDB+ and SocNav1, and analyse robustness under sparse preference feedback. Across datasets and backbones, StARS consistently improves performance and agreement with annotators, supporting personalised action selection aligned with user norms. Our code is publicly available at \href{https://github.com/Cambridge-AFAR/StARS.git}{https://github.com/Cambridge-AFAR/StARS.git}.

\end{abstract}

\section{INTRODUCTION}

\begin{figure}
    \centering
    \includegraphics[width=0.9\linewidth]{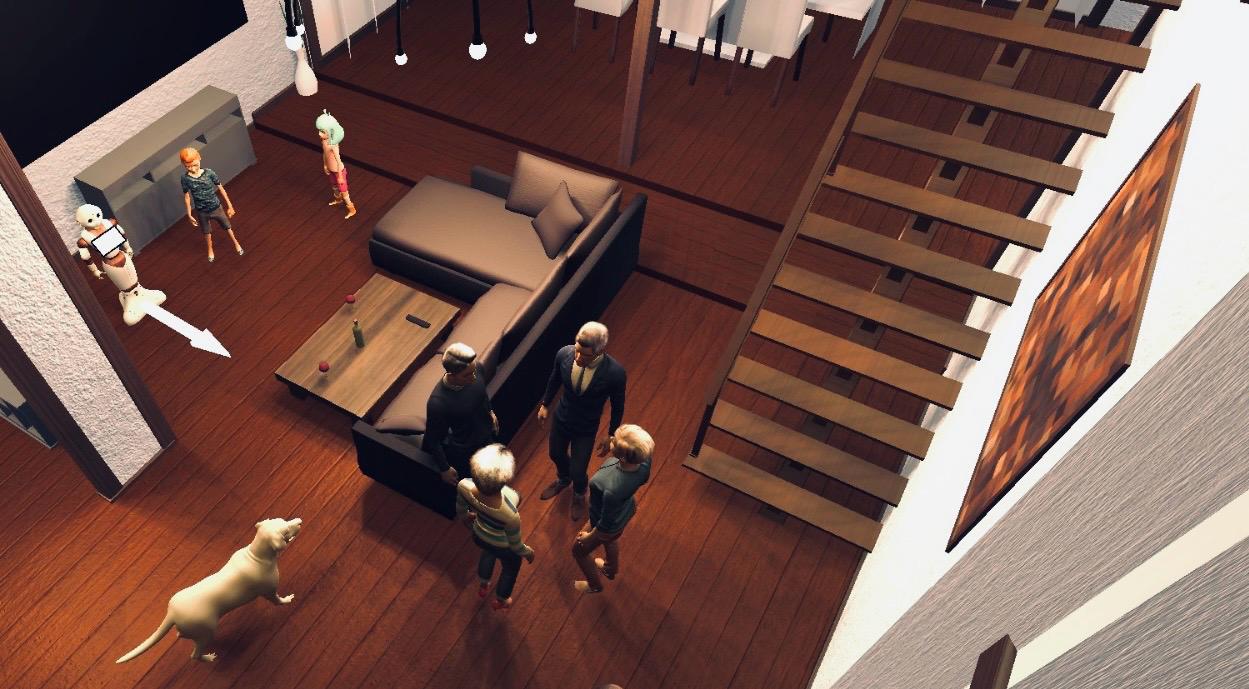}
    \caption{Social appropriateness depends on personalised preferences. In such a scene, one user may consider a robot serving food as appropriate, another might find it unsafe around an animal or children.}
    \vspace{-2em}
    \label{fig:motivate}
\end{figure}

Socially appropriate robot actions are central to human--robot interaction (HRI) because they influence how people interpret robots’ intentions, competence, and trustworthiness~\cite{christoforakos2021can, lyons2023explanations}. Importantly, social appropriateness is often dependent on users' personalised preferences: the same robot action in the same situation can be perceived as helpful by one person and intrusive or unsafe by another -- see Figure~\ref{fig:motivate} for an illustrative example. 
For instance, during a crowded social gathering, a user may expect a robot serving drinks, while others may find this distracting or risky. If a robot repeatedly violates a user’s expectations, people may disengage, resist assistance, or lose trust, even if the task is correctly executed~\cite{lawrence2025examining}. This makes personalisation essential for deploying robots in close-proximity settings such as homes, healthcare, education, and public spaces~\cite{manners,lasota2015analyzing}.

Recent work has addressed social appropriateness from several complementary angles. LLM-enhanced social robots~\cite{kim2024understanding, pinto2025designing} can produce context-aware responses by conditioning on rich natural-language descriptions of the interaction setting. Reinforcement-learning (RL)-based social robots~\cite{akalin2021reinforcement, qureshi2016robot} can be effective when immediate interaction feedback is available, enabling policies to adapt online~\cite{lin2020review}. Recent work~\cite{grace} also leverages user explanations to tailor robot actions to individual preferences, specifically for cases when people might not agree with one another. Together, these approaches show that social appropriateness can be improved by incorporating context, feedback, and user-facing adaptation mechanisms.

Despite these advances, a key challenge remains under-addressed: these approaches do not explicitly model the fact that \emph{different users may prefer different actions as socially appropriate under identical conditions}. In this work, we draw on recommender systems (RSs), which are designed to model user preferences and make personalised selections from many candidate options~\cite{lu2012recommender} (i.e., robot actions in this work). We formulate socially appropriate action generation as a preference estimation problem and ask the following research questions: whether integrating individual user/annotator preferences improves selection of socially appropriate robot actions across different model architectures and datasets (\textbf{RQ1}), and whether any observed differences are consistent and systematic rather than artefacts of a particular model or dataset (\textbf{RQ2}). Lastly, as collecting personalised preference feedback is costly and typically sparse~\cite{mizuchi2020optimization}, we ask whether the RS approach remains robust when each user/annotator provides only limited preference ratings (\textbf{RQ3}).

In this paper, we present StARS, a recommender-systems-based framework for socially appropriate robot action selection. For each robot action type, StARS builds an annotator--scene matrix, where each entry reflects how a specific annotator judges the appropriateness of that action in that scene. StARS then applies collaborative filtering (CF) via matrix factorisation (MF)~\cite{koren2009matrix, schafer2007collaborative}  %
to learn latent representations of annotator preferences and scene characteristics, enabling personalisation on top of backbone social-action models. We evaluate StARS on two socially aware robotics datasets (i.e., MannersDB+ and SocNav1~\cite{socnav1}) and show that our approach is model-agnostic, applicable to several model backbones. Our results also show that StARS provides systematic improvements across models and datasets, and remains effective under sparse preference feedback. Overall, our findings highlight the potential of recommender systems methods as a scalable approach to personalised decision-making in robotics and HRI.

\begin{figure*}[!t]
    \centering
    \includegraphics[width=0.9\textwidth]{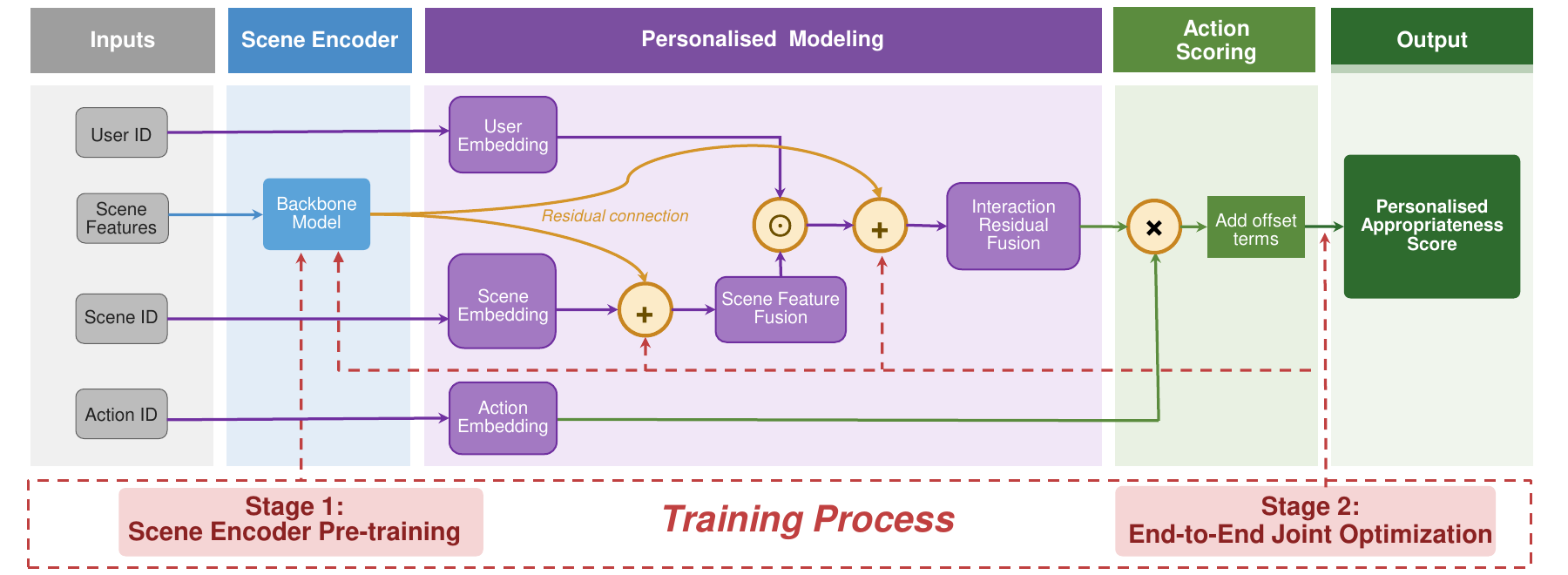} \vspace{-1mm}
    \caption{The overall pipeline of the StARS framework.}
    \label{fig:my_image}
    \vspace{-1.5em}
\end{figure*}

\section{Related Work}

\subsection{Socially Appropriate Robot Actions}

Social appropriateness is critical for HRI by shaping user perceptions~\cite{christoforakos2021can, lyons2023explanations}, and it has been widely studied across domains such as socially aware navigation~\cite{chen_socially_2017, tsoi_sean_2020, socnav1, karnan_socially_2022, gao_learning_2019, walters_robotic_2007}, and context-sensitive, norm-compliant action generation~\cite{grace, manners,churamani2024feature}. These works typically learn action suitability from human judgements, enabling robots to behave in ways that align with perceived social norms and contextual expectations~\cite{churamani2024feature, manners}. Datasets and benchmarks further support this line of research by providing annotated scenarios for evaluating socially acceptable robot behaviours~\cite{manners, socnav1}.

Many existing approaches still assume a single, shared notion of ``appropriate'' behaviour by relying on aggregated human ratings or shared representations of appropriateness~\cite{manners, churamani2024feature}. While such formulations enable stable learning and benchmarking, they often overlook systematic inter-individual differences in how robot actions are perceived~\cite{grace}. As a result, models trained on aggregated labels may smooth over meaningful disagreement, limiting their ability to account for diverse user expectations. Such limitations have motivated growing interest in explicit personalisation in HRI, enabling robots to adapt their behaviour to individual users’ preferences~\cite{hellou2021personalization}.

\subsection{Personalisation in HRI}
Prior work has highlighted personalisation as a key challenge for long-term HRI~\cite{irfan2019personalization}.  Early deployments showed that personalisation strategies can shape user experience over time~\cite{lee2012personalization}. Subsequent studies in socially assistive robotics demonstrated the importance of sustained, user-tailored adaptation in sensitive real-world settings, e.g., in-home interventions for children with autism spectrum disorders~\cite{clabaugh2019long}. %

Personalisation in HRI has been pursued primarily through interactive and feedback-driven paradigms. Socially aware reinforcement learning has been used to adapt robot behaviours from immediate user feedback, such as adjusting linguistic style in dialogue~\cite{ritschel2017adapting}. Complementary to RL, learning-from-demonstration methods capture user-specific preferences, including personalised human-aware navigation policies learned from virtual reality studies~\cite{de2022learning,de2023learning}. More recently, preference-driven representation learning has been proposed to support efficient adaptation without retraining models from scratch~\cite{wang2025personalization}.

While these approaches demonstrate promising directions for personalisation, many assume dense interaction histories, interactive querying, or immediate feedback~\cite{rossi2023preface, ritschel2017adapting, dai2024think}. These assumptions may not hold in many real-world HRI deployments, where only sparse and offline appropriateness judgements are available (e.g.,~\cite{manners}) and systematic inter-subject variability must still be captured. This motivates preference-modelling methods that can infer user-specific patterns from limited feedback by leveraging shared structure across users and contexts.

\subsection{Recommender Systems for Preference Modelling}

\Acfp{RS} have been used to predict user preferences and generate personalised item suggestions, effectively filtering relevant items from a large pool of candidates for individual users.
A typical RS task involves three elements: a set of \emph{users}, a set of \emph{items}, and a sparse set of observed interactions (e.g., ratings) indicating how much each user likes each item; the central goal is to infer the missing interactions.
One predominant technique in \acp{RS} over the past two decades is \acf{CF}~\cite{koren2021advances,aggarwal2016recommender}, which formulates recommendation as a ``matrix completion'' problem: given a sparse user-item rating matrix, the goal is to predict the missing ratings. 
Alternative approaches include the \acf{CB} approach, which leverages item attributes to match user profiles, and hybrid methods that integrate both CF and CB to exploit their complementary strengths~\cite{su2009survey,jannach2021recommender}.

The most widely adopted CF method is \acf{MF}~\cite{koren2009matrix,rendle2009bpr}, which factorises the user-item matrix into low-rank latent factors. 
With the advances of machine learning, deep neural networks have been incorporated to improve predictive performance, e.g., neural CF~\cite{he2017neural}.
However, their reported gains over simple CF baselines are often 
difficult to reproduce~\cite{ferrari2019we,canamares2018should}, 
suggesting that simple CF baselines, including MF, remain a strong and reliable choice. Therefore, we adopt CF as a principled way to model individual preference variation from sparse annotator feedback, and we focus on MF as a strong and stable foundation.

To date, only a very limited number of works have explicitly connected RSs and socially interactive systems. In particular, a recent work~\cite{huang2026reimagining} provides a conceptual framework, and it argues for integrating RS techniques as modular components to rank and select robot actions. From an HCI perspective, Swearingen and Sinha~\cite{swearingen2001beyond} note that effective recommender systems should support user trust and allow users to refine recommendations, which closely mirror requirements in HRI. Building on these insights, \textbf{this paper provides the first computational work that combines RS-based collaborative filtering with social robot action selection and empirically evaluates this formulation on socially aware robotics datasets}.  %

\section{Methodology}
\label{sec:method}

\subsection{Problem Formulation and General Framework}
\label{subsec:reformulation}
Let $\mathcal{D}$ be a set of collected annotations, where each annotation $(u,i,a,y_{u,i,a})\in\mathcal{D}$ records the appropriateness score $y_{u,i,a}$ that a user $u \in \mathcal{U}$ assigned to a candidate robot action $a \in \mathcal{A}$ in a scene $i \in \mathcal{I}$.
The traditional approach is to predict an aggregated appropriateness score $y_{i,a} \approx |\{u:(u,i,a)\in\mathcal{D}\}|^{-1}\sum_{u\,:\,(u,i,a)\in\mathcal{D}} y_{u,i,a}$ across different users, and subsequently select the most appropriate action based on these predicted scores.\footnote{For simplicity, we omit the score $y_{u,i,a}$ in set-membership expressions and write $(u,i,a)\in\mathcal{D}$ as shorthand for $(u,i,a,y_{u,i,a})\in\mathcal{D}$.}
However, such aggregation fails in practice, as users exhibit significant disagreement on appropriateness judgments~\cite{grace}, leading to information loss that aggregated scores cannot capture.
To address this, we refine the task as \textbf{personalised appropriateness score prediction}: to predict the appropriateness score $y_{u,i,a}$ that a user $u \in \mathcal{U}$ would assign to action $a$ in scene $i$. 
This formulation explicitly models inter-user disagreement and enables personalised action selection aligned with individual user norms, rather than relying solely on majority preference.
Our key perspective is that personalisation can be achieved by enhancing any differentiable appropriateness predictor with a CF module.

Formally, our goal is to train a system $\mathcal{S}(u, i, a; \Theta) 
\rightarrow \hat{y}_{u,i,a}$ that takes as input the user 
$u$, scene $i$, and action~$a$, outputting personalised 
appropriateness scores to enable action selection tailored 
to individual user preferences.

Concretely, within~$\mathcal{S}$, each scene instance $i$ is associated with a feature representation $\bm{x}_i$ derived from scene-side information.
Our framework is agnostic to how $\bm{x}_i$ is obtained and can take any scene-side representation as input. In our experiments, $\bm{x}_i$ is the pre-computed scene representation provided with MannersDB+, and, for SocNav1, the scene-graph representation in which each node (robot, human, wall, room, object, interaction) is encoded by a one-hot type and geometric features relative to the robot (distance and orientation)~\cite{socnav_graph}.

Given the encoded scene features $\bm x_i$, our objective is then to train a predictor $f(u, i, \bm{x}_i, a) \rightarrow \hat{y}_{u,i,a}$ that outputs the personalised appropriateness score for user $u$, scene $i$, and action $a$, by minimizing the following objective over the collected annotations $\mathcal{D}$:
\begin{equation}
    \mathcal{L} = \sum_{(u,i,a) \in \mathcal{D}} \delta(y_{u,i,a}, \hat{y}_{u,i,a}),
    \label{eq:mse}
\end{equation}%
\noindent where $\delta(\cdot, \cdot)$ is a task-specific loss function, \eg mean squared error (MSE).

\subsection{Personalised Appropriateness Score Prediction}
\label{subsec:mf}
We construct the personalised predictor $f$ by augmenting a differentiable \emph{scene encoder} $\phi_\theta$
with a collaborative-filtering (CF) module based on matrix factorization (MF), a widely used and reliable CF model.

Classic MF predicts a scalar rating via low-rank decomposition of a sparse user--item matrix.
In our setting, we predict individual appropriateness scores across multiple robot actions $a \in \mathcal{A}$.
A naive extension would build action-independent user--scene matrices (one per action), but this
is affected by data sparsity, since each action-specific subset $\mathcal{D}_a$ contains only a fraction of the
full annotations $\mathcal{D}$.
To mitigate sparsity, we adopt a \textbf{multi-task MF} formulation that shares user and scene factors across actions,
while using action-specific embeddings.

Specifically, we learn: (i) shared user embeddings $\bm{p}_u \in \mathbb{R}^d$,
(ii) shared scene embeddings $\bm{q}_i \in \mathbb{R}^d$,
(iii) user, scene, and global offset terms $b_u$, $b_i$, $b$ capturing systematic rating tendencies,
and (iv) action embeddings $\bm{r}_a \in \mathbb{R}^d$.

In its standard form, multi-task MF predicts:
\begin{equation}
\hat{y}_{u,i,a} = (\bm{p}_u \odot \bm{q}_i)^\top \bm{r}_a + b_u + b_i + b,
\label{eq:pred_base}
\end{equation}
where $\odot$ denotes the Hadamard product.

In order to improve a content-only baseline with collaborative signals, we augment this multi-task MF backbone
with scene-side content features via residual fusion.
Specifically, we use the content embedding $\phi_\theta(\bm{x}_i)$ to enrich the collaborative item embedding $\bm{q}_i$.

\vspace{1mm}
\noindent\textbf{Scene Residual Fusion.}
Pure MF relies only on user and scene identifiers.
To incorporate scene-side information, we use $\phi_\theta$ as a differentiable scene encoder. Here, scene-side information denotes contextual scene properties, such as
spatial layout, relative geometry, and interaction relations. We then define
a content embedding:
\begin{equation}
\bm{h}_i := \phi_\theta(\bm{x}_i) \in \mathbb{R}^d.
\label{eq:hi_def}
\end{equation}
We then form a hybrid scene embedding by residual fusion:
\vspace{-1em}
\begin{equation}
\bar{\bm{q}}_i := \bm{q}_i + \bm{h}_i,
\label{eq:scene_fusion}
\end{equation}%
\noindent which preserves collaborative signal in $\bm{q}_i$ while enhancing it with content information from $\bm{x}_i$.
Built on this, we further propose an improvement for the generalization under sparse annotations.

\vspace{1mm}
\noindent\textbf{Interaction Residual Fusion.} 
The user-scene interaction is modeled via element-wise product $\bm{p}_u \odot \bar{\bm{q}}_i$, which re-weights the hybrid scene representation according to user-specific preferences. To further inject scene content information at the prediction level, we additionally add $\bm{h}_i$ as a residual term, yielding the final prediction:

\vspace{-1em}
\begin{equation}
\hat{y}_{u,i,a} = 
\Bigl(\underbrace{\underbrace{\bm{p}_u \odot \bar{\bm{q}}_i}_{\text{personalised scene repr.}} + \bm{h}_i}_{\text{residual fusion}}\Bigr)^\top\, 
\underbrace{\bm{r}_a}_{\text{action emb.}} + 
\underbrace{b_u + b_i + b}_{\text{offset terms}}.
\label{eq:final}
\end{equation}

This additional interaction residual term provides a user-independent content shortcut that stabilizes prediction when collaborative factors are weakly estimated (sparse regime).

\subsection{Optimization}
\label{subsec:training}

Training the full system $\mathcal{S}$ uses a two-step strategy: we first pre-train the scene encoder $\phi_\theta$ with a temporary prediction head, and then jointly optimize the full personalised predictor $f$ end-to-end.

\vspace{1mm}
\noindent\textbf{Step 1: Scene encoder pre-training.} 
We first pre-train $\phi_\theta$ by attaching a temporary linear regression head $g_\psi$ on top to predict action appropriateness scores from scene features~$\bm{x}_i$:
\vspace{-1em}
\begin{equation}
\hat{\bm{y}}_{u,i} = g_\psi(\phi_\theta(\bm{x}_i)) \in \mathbb{R}^{|A|},
\end{equation}
where the predicted score for action $a$ is obtained by indexing the corresponding component (i.e., $\hat{y}_{u,i,a} = \hat{\bm{y}}_{u,i}[a]$).
The parameters of $\phi_\theta$ and $g_\psi$ are optimized via AdamW by 
minimizing the loss~Eq.~(\ref{eq:mse}) with gradient clipping~\cite{loshchilov2017decoupled}. 
The temporary head $g_\psi$ is discarded after this step.

\vspace{1mm}
\noindent\textbf{Step 2: End-to-End Joint Optimization.} 
The pre-trained $\phi_\theta$ is used as a feature projection network to produce $\bm{h}_i$ (see Eq.~(\ref{eq:hi_def})). 
All parameters $\{\bm{p}_u, \bm{q}_i, \bm{r}_a, b_u, b_i, b\}$ and parameters in $\phi_\theta$ are then jointly updated by minimizing the loss function Eq.~(\ref{eq:mse}) using AdamW with gradient clipping, with predictions following Eq.~(\ref{eq:final}).

The full training procedure is described in Algorithm~\ref{alg:training}.

\begin{algorithm}[!t]
\caption{Two-Step Training of the StARS Framework.}
\small
\label{alg:training}
\begin{algorithmic}[1]
\Require Training data $\mathcal{D}$, scene features $\{\bm{x}_i\}$, 
         embedding dimension $d$, number of epochs $T_1$, $T_2$
\Ensure Trained parameters $\{\bm{p}_u, \bm{q}_i, \bm{r}_a, b_u, b_i, b, \phi_\theta\}$

\vspace{1mm}
\State \textbf{// Target Normalisation}
\For{each action $a \in \mathcal{A}$}
    \State Compute $\mu_a$ and $\sigma_a$ from training split $\mathcal{D}$
    \State $y'_{u,i,a} \leftarrow (y_{u,i,a} - \mu_a) / \sigma_a$ for all $(u,i,a) \in \mathcal{D}$
\EndFor

\vspace{1mm}
\State \textbf{// Step 1: Scene Encoder Pre-training}
\State Initialize $\phi_\theta$ and temporary linear head $g_\psi$ randomly
\For{epoch $= 1, \dots, T_1$}
    \For{each $(u, i, a) \in \mathcal{D}$}
        \State $\hat{y}_{u,i,a} \leftarrow g_\psi(\phi_\theta(\bm{x}_i))$
        \State Compute loss $\mathcal{L} = \delta(y'_{u,i,a}, \hat{y}_{u,i,a})$
        \State Update $\{\theta,\psi\}$ via AdamW
    \EndFor
\EndFor
\State Discard temporary head $g_\psi$

\vspace{1mm}
\State \textbf{// Step 2: End-to-End Joint Optimization}
\State Initialize $\{\bm{p}_u, \bm{q}_i, \bm{r}_a, b_u, b_i, b\}$ randomly
\State Initialize $\phi_\theta$ from Step 1
\For{epoch $= 1, \dots, T_2$}
    \For{each $(u, i, a) \in \mathcal{D}$}
        \State $\bm{h}_i \leftarrow \phi_\theta(\bm{x}_i)$
        \State $\hat{y}_{u,i,a} \leftarrow$ (Eq.~\ref{eq:final})
        \State Compute loss $\mathcal{L} = \delta(y'_{u,i,a}, \hat{y}_{u,i,a})$
        \State Update $\{\bm{p}_u, \bm{q}_i, \bm{r}_a, b_u, b_i, b, \phi_\theta \}$ 
               via AdamW
    \EndFor
\EndFor
\end{algorithmic}

\end{algorithm}

\section{Evaluation}

\subsection{Datasets}

In our experiments, we used two human-annotated datasets, MannersDB+ and SocNav1~\cite{socnav1}, because they provide complementary testbeds for personalised social appropriateness modelling. Both datasets retain annotator identity information, which is essential for user-specific modelling and supporting personalised prediction. MannersDB+ covers task-level action appropriateness in domestic interaction scenarios, whereas SocNav1 focuses on socially compliant navigation. Together, they allow us to test whether StARS generalises across different HRI domains, scene representations, and annotation sparsity regimes. The dataset statistics are provided in Table~\ref{tab:dataset_stats_mannersdb_socnav}.

MannersDB+ is an extension of the MannersDB dataset~\cite{manners} that %
provides three robot embodiments (\emph{Nao}, \emph{Pepper}, and \emph{PR2}) and social appropriateness annotations for simulated living room scenarios across multiple tasks: \emph{vacuum cleaning, mopping the floor, carrying warm food, carrying cold food, carrying drinks, carrying small objects, carrying large objects, cleaning,} and \emph{starting a conversation}~--~see Figure~\ref{fig:motivate} for an example scene from the dataset. 
This makes MannersDB+ a suitable benchmark for learning preference/appropriateness models over robot actions. %

SocNav1 targets socially-aware navigation and provides human ratings of the relative suitability of navigation-relevant actions for a given scene~\cite{socnav1}. Unlike MannersDB+, which focuses on action appropriateness in domestic interaction scenarios, SocNav1 emphasises the appropriateness of a robot's navigation. %
The dataset is designed to support both benchmarking and supervised learning of socially compliant navigation behaviour, and it is widely used to evaluate models that estimate human comfort/discomfort or social acceptability in navigation scenarios  \cite{hatice_socnav, socnav_graph}.

\begin{table}[t!]
\centering
\small
\caption{Dataset statistics for MannersDB+ and SocNav1.} \vspace{-1mm}
\label{tab:dataset_stats_mannersdb_socnav}
\begin{tabular}{lcc}
\hline
\textbf{Metric} & \textbf{MannersDB+} & \textbf{SocNav1} \\
\hline
Total number of annotations          & 9238   & 9280 \\
Number of annotators                 & 444    & 12 \\
Average annotations per scene        & 3.08   & 1.89 \\
Average annotations per annotator    & 20.81  & 773.33 \\
Mean of scores (normalised)              & 0.58   & 0.57 \\
Stdev.\ of scores (normalised)          & 0.18   & 0.34 \\
\hline
\end{tabular}
\vspace{-1.2em}
\end{table}

\subsection{MannersDB+ Experiments}
In MannersDB+ experiments, we compared several backbone models that predict appropriateness directly from the scene/action representation: 
\textbf{MLP}, a standard feed-forward network that serves as a strong baseline for direct prediction from engineered features; \textbf{ResMLP}~\cite{resmlp-fttransformer}, which adds residual connections to improve optimization and stabilize deeper MLP stacks; \textbf{DCNv2}~\cite{dcnv2}, which combines an explicit cross network (for structured feature interactions) with a deep nonlinear tower, making it well suited to recommendation-style tabular prediction; \textbf{GraceAE}~\cite{grace}, GRACE-inspired autoencoder backbone, which learns a compact latent representation of scene--action inputs before prediction; and \textbf{FT-Transformer}~\cite{resmlp-fttransformer}, a Transformer-based tabular model that treats features as tokens and uses self-attention to capture context-dependent feature interactions. These backbones provide a diverse set of inductive biases for modelling social appropriateness, against which we evaluate the gains from adding personalised CF components.

\subsection{SocNav1 Experiments}
Prior work has mostly evaluated  SocNav1 using a range of graph neural network (GNN) backbones~\cite{socnav_graph}. Accordingly, in this work, 
we used GNN encoders based on message passing over scene graphs (e.g., robot, humans, walls, objects, and interactions), where node representations are iteratively updated using neighbourhood information. Specifically, our experiments included \textbf{GCN}~\cite{gcn}, which performs neighborhood aggregation with graph-convolutional smoothing and provides a strong baseline for relational scene encoding; \textbf{GIN}~\cite{gin}, which uses sum aggregation and MLP updates and is known for strong discriminative power; \textbf{GraphSAGE}~\cite{graphsage}, which learns inductive neighborhood aggregation functions and generalizes to unseen graphs; \textbf{GGNN}~\cite{ggnn}, which incorporates recurrent gated updates for multi-step relational reasoning; \textbf{RGCN}~\cite{rgcn}, which uses relation-specific message transformations to handle typed edges in multi-relational scene graphs; and \textbf{GAT}~\cite{gat}, which applies attention over neighbors so the model can weight socially salient entities more strongly during message aggregation.

\subsection{Training, Evaluation, and Implementation Details}

We evaluated models using five-fold cross-validation over each dataset. In each fold $k$, we  split the full set of annotated user--scene examples (where each row corresponds to an annotator's rating of a scene)
into a training pool $\mathcal{D}^{(k)}_{\mathrm{pool}}$ containing approximately $80\%$ of the rows and a held-out test set $\mathcal{D}^{(k)}_{\mathrm{test}}$ containing the remaining $20\%$.
To quantify data efficiency, for each training fraction $f$, we 
uniformly sampled without replacement a subset $\mathcal{D}^{(k)}_{\mathrm{train}}(f) \subset \mathcal{D}^{(k)}_{\mathrm{pool}}$ of size approximately $f \cdot |\mathcal{D}^{(k)}_{\mathrm{pool}}|$. We then trained the model on $\mathcal{D}^{(k)}_{\mathrm{train}}(f)$ and evaluated it on $\mathcal{D}^{(k)}_{\mathrm{test}}$. We aggregated results across the five folds and report macro-averaged performance across actions.

All experiments used a two-stage training protocol, with a 2-fold inner cross-validation loop for model selection. Hyperparameter selection in the inner loop was performed over a grid of epoch pairs. %
In the experiments, we evaluated the epoch grid $\{(10,100), (50,100), (100,100)\}$ and selected the best pair based on inner-validation RMSE on normalised scores. After selection, the model was retrained on the full outer-train subset using the chosen epoch pair.

We trained with AdamW optimisation, using a learning rate of $5\times10^{-5}$, weight decay of $1\times10^{-5}$, and gradient clipping. Within each training split, we additionally applied z-score normalisation using the split statistics, and invert this transformation at evaluation time.

\subsection{Metrics and Statistical Testing}
To assess RQ1 (comparison of base models and CF variants) and RQ3 (sensitivity to data sparsity), we reported performance with standard regression metrics, macro-averaged across actions. 
To quantify scale-dependent error, we used dataset-specific metrics to follow prior evaluation conventions and enable direct comparison with previous work: \textbf{RMSE} (lower the better) for MannersDB+, and \textbf{MSE} (lower the
better) for SocNav1 after normalising ratings to $[0,1]$, as done in prior
work~\cite{socnav_graph}.
Beyond these measures, we reported whether the trend in the predictions is aligned with the variation in user scores across samples.  
Specifically, we reported Pearson's correlation coefficient (\textbf{Pearson $r$}, higher the better) and Lin's concordance correlation coefficient (\textbf{CCC}, higher the better), consistent with prior work on socially appropriate action generation~\cite{churamani2024feature, grace, manners}.

To assess RQ2 (whether incorporating CF yields consistent differences over base models), we used a \textbf{ cross-validation--corrected paired $\textbf{t}$-test (Bouckaert--Frank correction~\cite{bouckaert2004evaluating})}. To control the familywise error rate arising from multiple comparisons, we adjusted the resulting two-sided $p$-values using Holm's method within each dataset and metric family (MannersDB+: $m=5$ backbones per metric; SocNav1: $m=6$ backbones per metric). We reported mean ($\Delta = \mathrm{CF} - \mathrm{Base}$), corrected $t$-statistics ($df=4$), and Holm-adjusted $p$-values. Negative $\Delta$ for RMSE indicates that CF reduces prediction error, whereas positive $\Delta$ for Pearson $r$ and CCC indicates improved agreement with the ground-truth ratings.

\section{Results}

\begin{table}[t!]
\centering
\caption{\small Five-fold cross-validation results on MannersDB+ and SocNav1. For each backbone, we report the scene-only model and its StARS variant.} \vspace{-1mm}
\label{tab:cv_results_combined}

\begin{subtable}[t]{0.49\textwidth}
\centering
\caption{\small RQ1 results on MannersDB+ (RMSE, Pearson $r$, and CCC).} \vspace{-1mm}
\footnotesize
\setlength{\tabcolsep}{3pt}
\renewcommand{\arraystretch}{1.15}
\begin{tabular}{lccc}
\hline
Model & RMSE $\downarrow$ & Pearson $r$ $\uparrow$ & CCC $\uparrow$ \\
\hline
MLP & $1.27 \pm 0.01$ & $0.30 \pm 0.01$ & $0.24 \pm 0.01$ \\
\textbf{MLP-StARS} & $\mathbf{1.13 \pm 0.01}$ & $\mathbf{0.53 \pm 0.01}$ & $\mathbf{0.48 \pm 0.01}$ \\
\hline
ResMLP \cite{resmlp-fttransformer} & $1.27 \pm 0.01$ & $0.30 \pm 0.02$ & $0.24 \pm 0.01$ \\
\textbf{ResMLP-StARS} & $\mathbf{1.12 \pm 0.01}$ & $\mathbf{0.54 \pm 0.02}$ & $\mathbf{0.48 \pm 0.02}$ \\
\hline
DCNv2 \cite{dcnv2} & $1.28 \pm 0.01$ & $0.30 \pm 0.01$ & $0.24 \pm 0.01$ \\
\textbf{DCNv2-StARS} & $\mathbf{1.13 \pm 0.01}$ & $\mathbf{0.53 \pm 0.02}$ & $\mathbf{0.48 \pm 0.01}$ \\
\hline
GraceAE \cite{denoising_ae, grace} & $1.31 \pm 0.01$ & $0.28 \pm 0.01$ & $0.24 \pm 0.01$ \\
\textbf{GraceAE-StARS} & $\mathbf{1.13 \pm 0.02}$ & $\mathbf{0.54 \pm 0.02}$ & $\mathbf{0.49 \pm 0.03}$ \\
\hline
FT-Transformer \cite{resmlp-fttransformer} & $1.26 \pm 0.01$ & $0.31 \pm 0.02$ & $0.25 \pm 0.01$ \\
\textbf{FT-Transformer-StARS} & $\mathbf{1.09 \pm 0.01}$ & $\mathbf{0.57 \pm 0.01}$ & $\mathbf{0.53 \pm 0.01}$ \\
\hline
\end{tabular}
\label{tab:cv_results_manners}
\end{subtable}

\vspace{0.8em}

\begin{subtable}[t]{0.49\textwidth}
\centering
\caption{\small  RQ1 results on SocNav1 (MSE, Pearson $r$, and CCC).} \vspace{-1mm}
\footnotesize
\setlength{\tabcolsep}{3pt}
\renewcommand{\arraystretch}{1.15}
\begin{tabular}{lccc}
\hline
Model & MSE $\downarrow$ & Pearson $r$ $\uparrow$ & CCC $\uparrow$ \\
\hline
GCN \cite{gcn} & $0.024 \pm 0.002$ & $0.90 \pm 0.00$ & $0.89 \pm 0.00$ \\
\textbf{GCN-StARS} & $\mathbf{0.020 \pm 0.001}$ & $\mathbf{0.92 \pm 0.00}$ & $\mathbf{0.91 \pm 0.00}$ \\
\hline
GIN \cite{gin} & $0.024 \pm 0.002$ & $0.90 \pm 0.00$ & $0.89 \pm 0.01$ \\
\textbf{GIN-StARS} & $\mathbf{0.018 \pm 0.001}$ & $\mathbf{0.92 \pm 0.00}$ & $\mathbf{0.92 \pm 0.01}$ \\
\hline
GraphSAGE \cite{graphsage} & $0.023 \pm 0.002$ & $0.91 \pm 0.00$ & $0.89 \pm 0.01$ \\
\textbf{GraphSAGE-StARS} & $\mathbf{0.016 \pm 0.001}$ & $\mathbf{0.93 \pm 0.01}$ & $\mathbf{0.93 \pm 0.01}$ \\
\hline
GGNN \cite{ggnn} & $0.020 \pm 0.001$ & $0.92 \pm 0.00$ & $0.91 \pm 0.01$ \\
\textbf{GGNN-StARS} & $\mathbf{0.014 \pm 0.001}$ & $\mathbf{0.94 \pm 0.00}$ & $\mathbf{0.94 \pm 0.00}$ \\
\hline
RGCN \cite{rgcn} & $0.019 \pm 0.001$ & $0.92 \pm 0.00$ & $0.91 \pm 0.00$ \\
\textbf{RGCN-StARS} & $\mathbf{0.015 \pm 0.000}$ & $\mathbf{0.94 \pm 0.00}$ & $\mathbf{0.93 \pm 0.00}$ \\
\hline
GAT \cite{gat} & $0.018 \pm 0.001$ & $0.92 \pm 0.00$ & $0.92 \pm 0.00$ \\
\textbf{GAT-StARS} & $\mathbf{0.013 \pm 0.000}$ & $\mathbf{0.94 \pm 0.00}$ & $\mathbf{0.94 \pm 0.00}$ \\
\hline
\end{tabular}
\label{tab:cv_results_socnav}
\end{subtable}

\vspace{-1em}

\end{table}

\subsection{RQ1: Model and Dataset Agnosticity}
\label{subsec:pa_rq1}
This section evaluates whether explicitly modelling individual annotator preferences via a recommender-systems formulation improves prediction accuracy over aggregate-label baselines. Table \ref{tab:cv_results_manners} presents the results for MannersDB+, adding CF leads to a substantial performance jump across all evaluated backbones (MLP, ResMLP, DCNv2, GraceAE, and FT-Transformer), with consistent reductions in RMSE and large gains in both Pearson $r$ and CCC. In particular, even the strongest baseline encoder (FT-Transformer) benefits from CF, which highlights that personalisation provides a complementary signal beyond improved modeling capacity.  %

As can be seen from Table \ref{tab:cv_results_socnav}, our methodology also extends well to SocNav1.  For every backbone (GCN, GIN, GraphSAGE, GGNN, RGCN, and GAT), the CF variant achieves lower error and higher correlation/concordance than its scene-only counterpart, indicating that the gains are robust to the choice of GNN architecture rather than being driven by a single model design. 
Overall, results from both datasets highlight that modelling user/annotator preferences via the collaborative-filtering (CF) layer yields better socially appropriateness predictions across different datasets and backbones, reflected by lower error and higher agreement.%

\begin{table}[t!]
\centering
\caption{\small Paired statistical comparison between Base models and StARS variants across folds. For each model, we report the mean paired difference $\Delta=\text{CF}-\text{Base}$ and test whether the mean difference differs from zero using a cross-validation--corrected paired $t$-test (Bouckaert--Frank correction; df$=4$). Two-sided $p$-values are Holm-adjusted within each metric.} %
\vspace{-1mm}
\label{tab:paired_ttest_combined}

\begin{subtable}[t]{0.49\textwidth}
\centering
\caption{\small RQ2 results on MannersDB+ ($m=5$).} \vspace{-1mm}
\small
\setlength{\tabcolsep}{4pt}
\resizebox{\textwidth}{!}{%
\begin{tabular}{llc|cc}
&  & Difference ($\Delta$) & $t(4)$ & $p$ (two-sided; Holm) \\
\hline
\multirow{5}{*}{\textbf{RMSE} $\downarrow$}
& MLP            & $-0.15$ & $-23.14$ & $1.03\times 10^{-4}$ \\
& ResMLP         & $-0.14$ & $-16.66$ & $1.26\times 10^{-4}$ \\
& DCNv2          & $-0.15$ & $-17.46$ & $1.26\times 10^{-4}$ \\
& GraceAE      & $-0.18$ & $-22.04$ & $1.03\times 10^{-4}$ \\
& FT-Transformer & $-0.17$ & $-20.78$ & $1.03\times 10^{-4}$ \\
\hline
\multirow{5}{*}{\textbf{Pearson $r$} $\uparrow$}
& MLP            & $+0.24$ & $56.51$ & $2.94\times 10^{-6}$ \\
& ResMLP         & $+0.23$ & $31.49$ & $1.82\times 10^{-5}$ \\
& DCNv2          & $+0.24$ & $29.85$ & $1.82\times 10^{-5}$ \\
& GraceAE       & $+0.25$ & $24.92$ & $1.82\times 10^{-5}$ \\
& FT-Transformer & $+0.26$ & $34.21$ & $1.74\times 10^{-5}$ \\
\hline
\multirow{5}{*}{\textbf{CCC} $\uparrow$}
& MLP            & $+0.24$ & $117.32$ & $1.58\times 10^{-7}$ \\
& ResMLP         & $+0.24$ & $23.52$  & $3.87\times 10^{-5}$ \\
& DCNv2          & $+0.24$ & $31.69$  & $1.77\times 10^{-5}$ \\
& GraceAE       & $+0.25$ & $16.67$  & $7.59\times 10^{-5}$ \\
& FT-Transformer & $+0.28$ & $38.93$  & $1.04\times 10^{-5}$ \\
\hline
\end{tabular}%
}
\label{tab:paired_ttest_manners}
\end{subtable}

\vspace{0.8em}

\begin{subtable}[t]{0.49\textwidth}
\centering
\caption{\small RQ2 results on SocNav1 ($m=6$).} \vspace{-1mm}
\small
\setlength{\tabcolsep}{4pt}
\resizebox{\textwidth}{!}{%
\begin{tabular}{llc|cc}
&  & Difference ($\Delta$) & $t(4)$ & $p$ (two-sided; Holm) \\
\hline
\multirow{6}{*}{\textbf{MSE} $\downarrow$}
& GCN       & $-4.34\times 10^{-3}$ & $-5.00$  & $2.24\times 10^{-2}$ \\
& GIN       & $-6.01\times 10^{-3}$ & $-4.53$  & $2.24\times 10^{-2}$ \\
& GraphSAGE & $-6.51\times 10^{-3}$ & $-4.39$  & $2.24\times 10^{-2}$ \\
& GGNN      & $-5.91\times 10^{-3}$ & $-6.07$  & $1.49\times 10^{-2}$ \\
& RGCN      & $-4.86\times 10^{-3}$ & $-8.24$  & $5.90\times 10^{-3}$ \\
& GAT       & $-4.40\times 10^{-3}$ & $-23.46$ & $1.17\times 10^{-4}$ \\
\hline
\multirow{6}{*}{\textbf{Pearson $r$} $\uparrow$}
& GCN       & $+1.94\times 10^{-2}$ & $11.24$ & $1.43\times 10^{-3}$ \\
& GIN       & $+2.17\times 10^{-2}$ & $10.35$ & $1.48\times 10^{-3}$ \\
& GraphSAGE & $+2.39\times 10^{-2}$ & $7.90$  & $2.00\times 10^{-3}$ \\
& GGNN      & $+2.46\times 10^{-2}$ & $8.61$  & $2.00\times 10^{-3}$ \\
& RGCN      & $+2.19\times 10^{-2}$ & $12.08$ & $1.35\times 10^{-3}$ \\
& GAT       & $+1.90\times 10^{-2}$ & $17.22$ & $4.00\times 10^{-4}$ \\
\hline
\multirow{6}{*}{\textbf{CCC} $\uparrow$}
& GCN       & $+2.14\times 10^{-2}$ & $9.48$  & $3.46\times 10^{-3}$ \\
& GIN       & $+3.02\times 10^{-2}$ & $3.82$  & $2.97\times 10^{-2}$ \\
& GraphSAGE & $+3.31\times 10^{-2}$ & $4.10$  & $2.97\times 10^{-2}$ \\
& GGNN      & $+2.91\times 10^{-2}$ & $7.12$  & $6.17\times 10^{-3}$ \\
& RGCN      & $+2.49\times 10^{-2}$ & $7.79$  & $5.85\times 10^{-3}$ \\
& GAT       & $+2.15\times 10^{-2}$ & $17.78$ & $3.53\times 10^{-4}$ \\
\hline
\end{tabular}%
}
\label{tab:paired_ttest_socnav}
\end{subtable}

\vspace{-2.5em}

\end{table}

\subsection{RQ2: Improvement Consistency and Systematicity}

\label{subsec:pa_rq2}
This section tests whether adding collaborative filtering (CF) consistently improves performance across different
scene encoders/backbones and across datasets, beyond a single chosen architecture. Table~\ref{tab:paired_ttest_manners} quantifies the reliability of the CF improvements on MannersDB+. %
Across all evaluated backbones, CF yields statistically significant gains: RMSE decreases by roughly $0.15$, while both correlation metrics increase substantially. The uniformly small $p$-values (down to $10^{-7}$) suggest that these improvements are systematic rather than an artefact of fold-level variation.

In Table \ref{tab:paired_ttest_socnav}, we report the comparison %
on the SocNav1 dataset. %
As also seen earlier on MannersDB+, the performance increase is statistically significant for all backbones.  Although the absolute gains on SocNav1 are smaller, the paired $t$-statistics are large in magnitude, indicating that the improvements are consistent in direction and reliable across folds. %
Overall, results from both datasets show the benefit of incorporating CF is consistent and systematic: improvements are statistically significant and persist across backbones, indicating the effect is not specific to a single architecture, dataset, or split.

\begin{figure*}[t!]
    \centering

    \begin{subfigure}[t]{0.48\linewidth}
        \centering
        \includegraphics[width=0.9\linewidth]{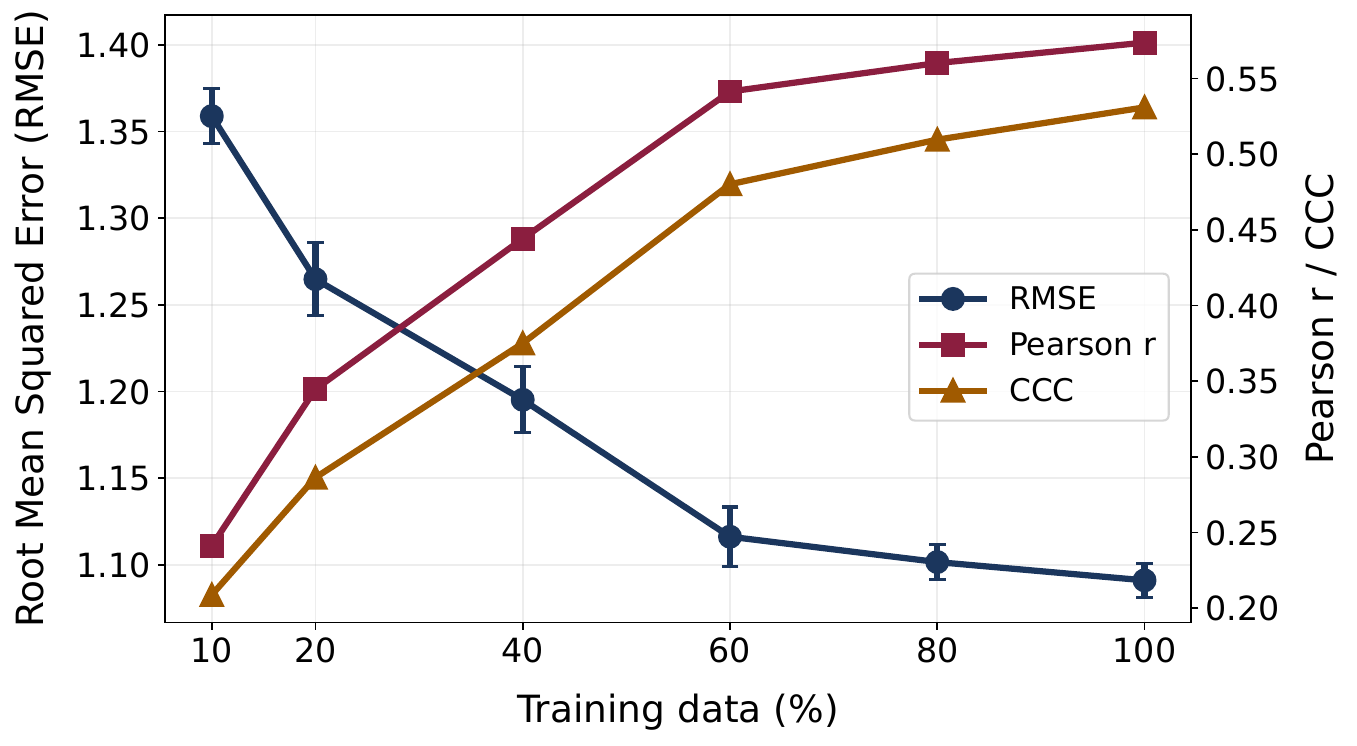}
        \vspace{-0.5em}
        \caption{RQ3 results using FT-Transformer-StARS on MannersDB+.}
        \label{fig:kfold_sweep}
    \end{subfigure}%
    \hfill%
    \begin{subfigure}[t]{0.48\linewidth}
        \centering
        \includegraphics[width=0.9\linewidth]{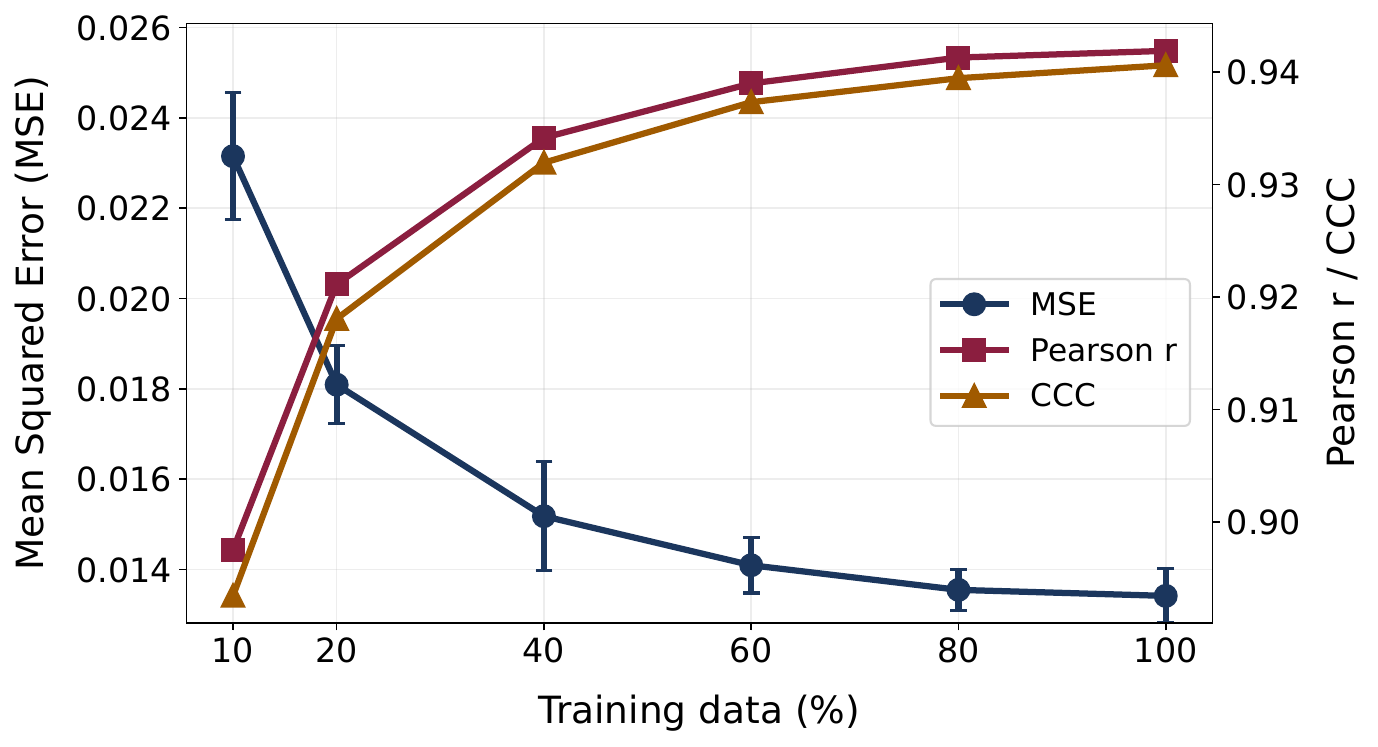}
        \vspace{-0.5em}
        \caption{RQ3 results using GAT-StARS on SocNav1.}
        \label{fig:gat_sweep}
    \end{subfigure}%
    \vspace{-0.5em}

    \caption{\small Scaling behavior with increasing training data. Macro-averaged RMSE/MSE (five-fold mean) and agreement metrics (Pearson’s $r$ and CCC) are plotted against the fraction of the training dataset used.}
    \label{fig:model_sweeps}

    \vspace{-2em}

\end{figure*}

\subsection{RQ3: Robustness to Data Sparsity}
\label{subsec:pa_rq3}
This section studies data-efficiency and sparsity robustness by varying the fraction of available training interactions.
We expect the RS formulation to be particularly beneficial in low-data regimes where each user provides few ratings. 
 Figure~\ref{fig:kfold_sweep} reports a learning-curve on MannersDB+ for the FT-Transformer-StARS model, where performance is aggregated across folds and plotted against the fraction of the full dataset used for training. As training data increases, error (RMSE) decreases while correlation metrics (Pearson’s $r$ and CCC) increase, indicating steadily improving predictive accuracy and correlation with human judgments. 
 Notably, performance improves most rapidly in the sparse-data regime, indicating that early annotations provide substantial benefits.

Figure \ref{fig:gat_sweep} reports a similar trend for the GAT-StARS model on SocNav1. As observed on MannersDB+, our approach is particularly effective in the low-data regime, where user--scene interaction signals are sparse, and the personalised latent factors are still weakly constrained. As the training fraction approaches the full dataset, the curves show diminishing returns, suggesting that once user preferences and scene representations are estimated more reliably, additional ratings yield smaller incremental improvements, and performance becomes increasingly bounded by the intrinsic ambiguity of social appropriateness judgments.
Overall, the results from both datasets show robust performances under sparse feedback: substantial performance improvements are achieved with early annotations, and additional feedback continues to refine predictions.

\section{Discussion and Conclusion}

In this work, we presented StARS, a recommender-systems-based framework for selecting socially appropriate robot actions by reformulating it as a preference estimation problem over annotator--scene interactions. The framework integrates collaborative filtering (CF) with a scene encoder to model both annotator preferences and scene characteristics. Across two social appropriateness datasets and multiple model backbones, our results show that adding the RS component improves performance and yields appropriateness scores that better align with human judgements.

Regarding \textbf{RQ1} and \textbf{RQ2}, our results show that StARS is model-agnostic and yields systematic, consistent improvements across settings. Specifically, integrating individual annotator preferences improves the selection of socially appropriate robot actions across different model architectures and datasets. This is important for robotics because it enables user-specific action selection without redesigning the underlying perception or decision-making backbone, supporting more adaptive behaviour in real-world HRI. Further, CF-based personalisation significantly reduces prediction error and improves agreement with human ratings (e.g., Pearson $r$ and CCC), indicating that variation in social appropriateness judgements is structured and can be captured through user--scene interaction modelling. More broadly, these findings suggest that recommender-systems techniques can serve as a practical personalisation layer in HRI, supporting robots that adapt their action choices to different users instead of relying on a single ``average'' notion of appropriateness.

For \textbf{RQ3}, we found that the RS formulation remains robust when preference feedback is limited and sparse. Learning-curve experiments show that StARS provides clear gains even at low data fractions, while additional annotator--scene ratings further improve performance. This data efficiency is critical for personalisation in HRI because it implies that a small number of early interactions can already produce meaningful user-specific preference estimates, creating a suitable environment for iterative personalisation as more feedback is obtained.  This is particularly important for robots deployed in the real world because collecting personalised feedback is costly and time-consuming~\cite{mizuchi2020optimization}, and robots often must adapt from a small amount of user-specific data~\cite{de2025towards}. Our results, therefore, support RS-based modelling as a data-efficient path toward personalised socially appropriate action generation.

Overall, this paper demonstrates the applicability of recommender systems methods for personalised interaction in robotics and HRI. Future work can extend StARS beyond social appropriateness; similar RS formulations could support other personalisation-centric HRI tasks. For example, in a collaborative robot assembly task, an RS-based personalisation layer could learn a user's preferred pace and level of assistance and rank alternative support strategies accordingly. 
Another important direction is real-time deployment in physical robots, where explicit appropriateness annotations may not be available. In such settings, the model could be initialised from offline annotations and then gradually adapt user-specific embeddings using implicit feedback, such as user corrections or task interruptions. %
Finally, future work could incorporate richer multimodal signals---including vision and language context, as well as non-verbal and affective user cues---to better infer user preferences and interaction state, enabling more context-sensitive and personalised robot behaviour.

\bibliographystyle{IEEEtranS}
\balance\bibliography{huang25_short}

\end{document}